\def\year{2022}\relax
\documentclass[letterpaper]{article} % DO NOT CHANGE THIS
\usepackage{aaai22}  % DO NOT CHANGE THIS
\usepackage{times}  % DO NOT CHANGE THIS
\usepackage{helvet}  % DO NOT CHANGE THIS
\usepackage{courier}  % DO NOT CHANGE THIS
\usepackage[hyphens]{url}  % DO NOT CHANGE THIS
\usepackage{graphicx} % DO NOT CHANGE THIS
\usepackage{natbib}  % DO NOT CHANGE THIS AND DO NOT ADD ANY OPTIONS TO IT
\usepackage{caption} % DO NOT CHANGE THIS AND DO NOT ADD ANY OPTIONS TO IT
\DeclareCaptionStyle{ruled}{labelfont=normalfont,labelsep=colon,strut=off} % DO NOT CHANGE THIS
\usepackage{algorithm}
\usepackage{algorithmic}

\usepackage{newfloat}
\usepackage{listings}
\floatstyle{ruled}
\newfloat{listing}{tb}{lst}{}
\floatname{listing}{Listing}
\usepackage[utf8]{inputenc} % allow utf-8 input
\usepackage{booktabs}       % professional-quality tables
\usepackage{amsfonts}       % blackboard math symbols
\usepackage{nicefrac}       % compact symbols for 1/2, etc.
\usepackage{xcolor}         % colors
\usepackage{multirow}
\usepackage{makecell}
\usepackage{mathtools}
\usepackage{lipsum}
\usepackage[flushleft]{threeparttable}
\usepackage{kotex}
\usepackage{adjustbox}

\title{SGRAM: Improving Scene Graph Parsing via Abstract Meaning Representation}
\author{
    Woo Suk Choi\textsuperscript{\rm 1}, Yu-Jung Heo\textsuperscript{\rm 1,3} and Byoung-Tak Zhang\textsuperscript{\rm 1,2}
}
\affiliations{
    \textsuperscript{\rm 1} Seoul National University
    \textsuperscript{\rm 2} AI Institute (AIIS), Seoul National University
    \textsuperscript{\rm 3} Surromind\\
    \{wschoi, yjheo, btzhang\}@bi.snu.ac.kr
}

\usepackage{bibentry}
\begin{document}

\maketitle

\begin{abstract}
Scene graph is structured semantic representation that can be modeled as a form of graph from images and texts. Image-based scene graph generation research has been actively conducted until recently, whereas text-based scene graph generation research has not. In this paper, we focus on the problem of scene graph parsing from textual description of a visual scene. The core idea is to use abstract meaning representation (AMR) instead of the dependency parsing mainly used in previous studies. AMR is a graph-based semantic formalism of natural language which abstracts concepts of words in a sentence contrary to the dependency parsing which considers dependency relationships on all words in a sentence. To this end, we design a simple yet effective two-stage scene graph parsing framework utilizing abstract meaning representation, SGRAM (Scene GRaph parsing via Abstract Meaning representation): 1) transforming a textual description of an image into an AMR graph (Text-to-AMR) and 2) encoding the AMR graph into a Transformer-based language model to generate a scene graph (AMR-to-SG). Experimental results show the scene graphs generated by our framework outperforms the dependency parsing-based model by 11.61\% and the previous state-of-the-art model using a pre-trained Transformer language model by 3.78\%. Furthermore, we apply SGRAM to image retrieval task which is one of downstream tasks for scene graph, and confirm the effectiveness of scene graphs generated by our framework.
\end{abstract}

\section{Introduction}
Understanding and reasoning a visual scene are simple tasks for human, but an AI system requires various techniques to implement them. One such technique is scene graph \cite{Johnson_2015_CVPR}, which is a graph-structured representation that captures high-level semantics of visual scenes (i.e. images) by explicitly modeling objects along with their attributes and relationships with other objects. Scene graph has proven effective in a variety of tasks, including image captioning \cite{Nguyen2021InDO, zhong2020comprehensive}, image manipulation \cite{Dhamo2020SemanticIM}, semantic image retrieval \cite{scenegraph_image_text_retrieve, Schroeder2020StructuredQI}, and visual question answering \cite{Koner2021GraphhopperMS, Damodaran2021UnderstandingTR}.

\begin{figure}[t]
\centering
    \includegraphics[width=0.9\columnwidth]{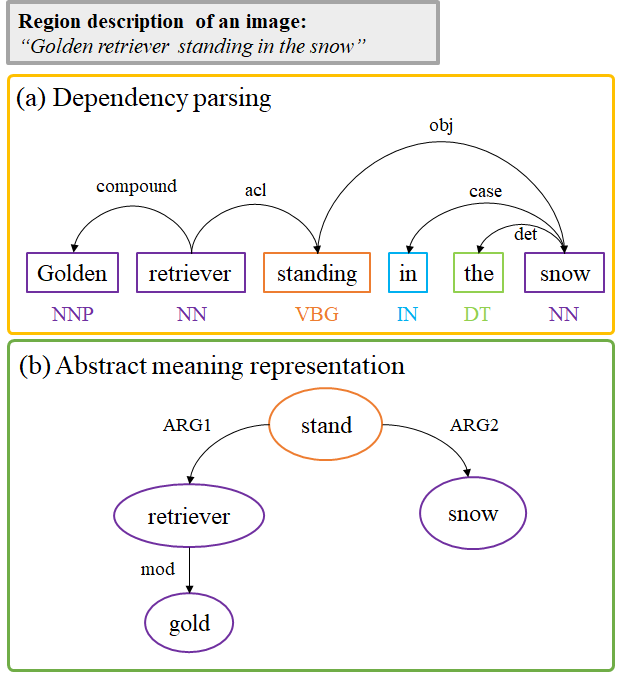}
\caption{An example of (a) dependency parsing and (b) abstract meaning representation (AMR) from the textual description (i.e. region description) of an image, \textit{``Golden retriever standing in the snow"}.}
\label{fig:fig1_rev1}
\end{figure}

Approaches for scene graph generation are classified into two categories: 1) scene graph generation based on image as input and 2) scene graph generation based on text (i.e. textual descriptions of an image) as input. The former approaches \cite{zellers2018scenegraphs, Gu_2019_CVPR, zhong2021SGGfromNLS, Lyu_2022_CVPR} learn and infer about visual representations through pre-trained object detectors (i.e. Faster R-CNN \cite{NIPS2015_14bfa6bb}) with category labels of object and predicate. The latter approaches \cite{schuster-etal-2015-generating, anderson_spice_ECCV, wang-etal-2018-scene, sgp_martin_neurips2019} infer by analyzing sentence structure and identifying relationships between individual words within a sentence in a direction of semantic dependency or expression. Most of studies have focused more on the former approach rather than the latter. The former approach generates scene graphs based on raw images, but it is easy to generate biased scene graphs by learning in supervised learning manner on limited category labels that follow a long-tailed distribution. On the other hand, the latter approach can handle various category labels of object, attribute, and predicate without restriction by applying semantic parsing technique (i.e. dependency parsing) to textual descriptions described for images. 

In this paper, we focus on the text-based scene graph generation approach, which is also called textual scene graph parsing. Most of previous works on textual scene graph parsing generate scene graphs using dependency parsing method to acquire dependency relationships for all words in a description. However, as shown in Figure \ref{fig:fig1_rev1} (a), considering the dependency relationships of all words in a description is unnecessary and may interfere with obtaining semantic information, since scene graph consists only of objects along with their attributes and relationships with other objects. Hence, apart from dependency parsing, we utilize abstract meaning representation (AMR) \cite{banarescu-etal-2013-abstract} to generate scene graph from textual descriptions of an image. As shown in Figure \ref{fig:fig1_rev1} (b), AMR is a popular formalism of natural language that abstracts concepts as semantic graphs from words of a sentence, and we therefore consider AMR is more suitable for scene graph parsing.

Thus, we design SGRAM, a simple yet effective two-stage scene graph parsing framework via abstract meaning representation. In the first stage, we use AMR parsing models to convert a textual description (i.e. region description) of an image into an AMR graph, and utilize various graph linearization techniques to linearize the AMR graph. On the second stage, we extend a pre-trained Transformer \cite{NIPS2017_3f5ee243}-based language model to generate scene graphs for the AMR graphs. We evaluate our framework on intersection of Visual Genome and MS COCO datasets. As a quantitative analysis of proposed framework, our framework shows significant improvement on scene graph parsing performance over the previous state-of-the-art model. Qualitatively, we visualize the comparison of scene graphs generated in our framework using various graph linearization techniques and the previous state-of-the-art model. Upon acceptance, we will release our code, which includes preprocessing data with AMR graphs as well as the models used.

The remainder of the paper is organized as follows. As related work, we describe the basic works of scene graph parsing, abstract meaning representation, and graph traversal algorithms, which are related to our framework. Then, we introduce Scene GRaph parsing via Abstract Meaning representation framework, SGRAM, in details. Next, we quantitatively and qualitatively demonstrate experiment results with implementation details.

\section{Related Work}
In this section, we first summarize scene graph parsing as we consider the problem of scene graph parsing. Then, Abstract Meaning Representation (AMR) is described. Lastly, we introduce graph traversal algorithms which will be used to linearize AMR graphs from textual descriptions of images. 

\subsection{Scene Graph Parsing}

Scene graph \cite{Johnson_2015_CVPR} is a graph-structured representation that represents rich structured semantics of visual scenes (i.e. images). Nodes in the scene graph represent either an object, an attribute for an object, or a relationship between objects. Edges depict the connection between two nodes. Scene graphs are obtained either from images \cite{zellers2018scenegraphs, Gu_2019_CVPR, zhong2021SGGfromNLS, Lyu_2022_CVPR} or textual descriptions of images \cite{schuster-etal-2015-generating, anderson_spice_ECCV, wang-etal-2018-scene, sgp_martin_neurips2019}. In this subsection, we introduce the studies of textual scene graph parsing since our work considers the problem of scene graph parsing from textual descriptions of an image. Most of the previous studies \cite{schuster-etal-2015-generating, anderson_spice_ECCV, wang-etal-2018-scene} used dependency parsing as a common theme. \cite{schuster-etal-2015-generating} proposed a rule-based and a learned classifier with dependency parsing. \cite{wang-etal-2018-scene} proposed a customized dependency parser with end-to-end training to parse scene graph, by extending the existing transition-based dependency parser \cite{kiperwasser-goldberg-2016-simple} to accommodate more sophisticated transition scheme. \cite{sgp_martin_neurips2019} proposed a customized attention graph mechanism using the OpenAI Transformer \cite{Radford2018ImprovingLU}\footnote{This model consists of a Byte-Pair-Encoding subword embedding layer followed by 12-layers of decoder-only transformer with masked self-attention heads.}. Previous works also made an attempt to solve scene graph parsing task utilizing AMR. However, the performance of AMR was not competitive compared to dependency parsers at the time.

To this end, we utilize the AMR to parse scene graphs and demonstrate better qualitative and quantitative performance.

\begin{figure*}[t]
\centering
    \includegraphics[width=1.\textwidth]{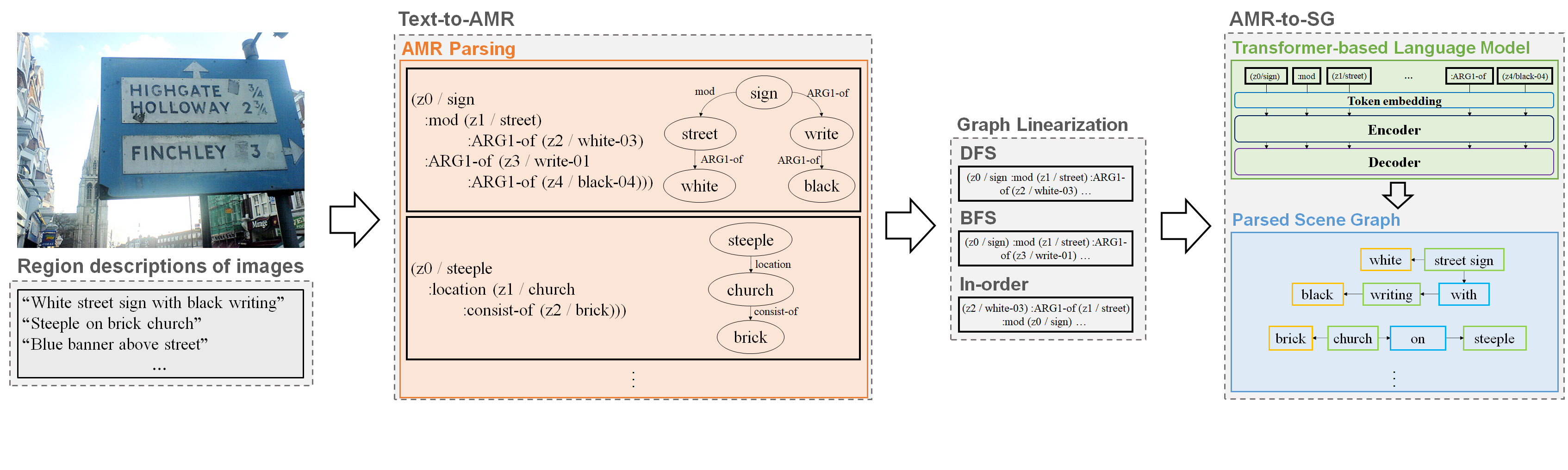}
\caption{The overall framework of Scene GRaph parsing via Abstract Meaning representation (SGRAM) for textual scene graph parsing task. Text-to-AMR illustrates transforming region descriptions of image into AMR graph. Graph linearization illustrates linearizing AMR graphs with various graph linearization techniques (DFS, BFS, In-order) based on graph traversal algorithms. AMR-to-SG illustrates parsing scene graphs from linearized AMR graphs using Transformer-based language model.}
\label{fig:model_figure}
\end{figure*}

\subsection{Abstract Meaning Representation}

Abstract meaning representation (AMR) \cite{banarescu-etal-2013-abstract} is a graph-based semantic representation which captures semantics ``\textit{who is doing what to whom}" in a sentence. Each sentence is represented as a rooted, directed, acyclic graph with labels on nodes (e.g. semantic concepts) and edges (e.g. semantic relations). The problem of dealing with AMR falls into two categories: \textit{Text-to-AMR} capturing the meaning of a sentence within a semantic graph and \textit{AMR-to-Text} generating a sentence from such a graph. The datasets used to handle these tasks are AMR2.0 (LDC2017T10) and AMR3.0 (LDC2020T02), which contain semantic treebanks of over $39,260$ and $59,255$ English natural language sentences, respectively, from broadcast conversations, weblogs, and web discussion forums. 

With the development of deep learning, neural models such as graph-to-sequence \cite{zhu-etal-2019-modeling}, sequence-to-graph \cite{cai-lam-2020-amr}, and neural transition-based parser \cite{zhou-etal-2021-amr} models have proposed. Furthermore, with the advent of Transformer-based language models, the studies \cite{Ribeiro2021InvestigatingPL, bevilacqua-etal-2021-one} incorporating the generation capability of pre-trained Transformer-based language models have proposed and shown impressive performances on AMR tasks lately. In particular, \cite{bevilacqua-etal-2021-one} has proposed a simple, symmetric approach for performing state-of-the art on AMR tasks with a single sequence-to-sequence architecture. We adopt this AMR parsing model \cite{bevilacqua-etal-2021-one} to generate AMR graph from textual descriptions of images.

\subsection{Graph Traversal Algorithms}

Graph traversal is a process of visiting all nodes once, in turn, starting with one node in a graph structure. Such traversal falls into two algorithms based on the order in which nodes are visited: Depth-First Search (DFS) and Breadth-First Search (BFS). These graph traversal algorithms have been used for graph linearization to generate from sentences to AMR graphs or vice versa in previous AMR studies \cite{konstas-etal-2017-neural, cai-lam-2019-core, bevilacqua-etal-2021-one}. \cite{konstas-etal-2017-neural} has used DFS including backward traversing to linearize an AMR graph, and put it into stacked-LSTM sequence-to-sequence neural architecture model to generate AMR graph. \cite{cai-lam-2019-core} has proposed AMR parsing method which constructs a parse graph incrementally in a top-down fashion with core-semantic first principle based on BFS. \cite{bevilacqua-etal-2021-one} has proposed SPRING model which uses DFS and BFS to transform an AMR graph into a sequence of symbols with special tokens and generates either a linearized AMR graph or sentence using a pre-trained language model. 

Besides these two algorithms, there is a special case of graph traversal called tree traversal, which refers to a process of visiting (e.g. retrieving, updating, or deleting) each node in a tree data structure. Three basic traversal algorithms in tree traversal are: pre-order, post-order, and in-order. 
The previous work \cite{Zhang2019AMRPA} for AMR pasring has used pre-order and in-order tree traversal algorithms to sort and linearize nodes of AMR graphs, and pre-order performed better than in-order. In this work, we take two graph traversal (DFS, BFS) and one tree traversal (in-order) algorithms to linearize AMR graphs and evaluate how much graph linearization with each graph traversal algorithm for AMR graph affects scene graph parsing.

\section{Methodology}

In this section, we describe Scene GRaph parsing via Abstract Meaning representation (SGRAM), two-stage scene graph parsing framework using abstract meaning representation with stage by stage. As shown in Figure \ref{fig:model_figure}, we first describe converting textual description (i.e. region descriptions) of an image to AMR graphs (Text-to-AMR), and explain linearizing the AMR graphs with various graph traversal algorithms such as DFS, BFS, and in-order (Graph Linearization). Then, we describe what kind of Transformer-based language model is used and how scene graphs are generated from the linearized AMR graphs (AMR-to-SG).

\subsection{Text-to-AMR}
We base the first stage of our framework on the AMR models of \cite{bevilacqua-etal-2021-one} called SPRING\footnote{https://github.com/SapienzaNLP/spring}. The SPRING model is a simple seq2seq model, which can handle to either parse an AMR graph from a given sentence or generate a sentence from a linearized AMR graph. The SPRING model adopts a large pre-trained encoder-decoder model, i.e. BART \cite{lewis-etal-2020-bart} with modifying BART vocabulary in order to make suitable for relations (e.g. ``\textit{:direction}", ''\textit{:mod}"), AMR-specific concepts (e.g ``\textit{:world-region}", ``\textit{date-entity}"), and frames (e.g. ``\textit{:ARG1}"). In addition, the SPRING model is experimented with different fully graph-isomorphic linearization techniques, customized DFS and BFS, and the linearization that work best was the one based on DFS with proposed special tokens \textit{z0}, {z1}, ...,\textit{$z_n$} indicating co-referring nodes. For our framework, we leverage pre-trained SPRING model on AMR 2.0 (LDC2017T10) and AMR3.0 (LDC2020T02) datasets to generate AMR graphs from region descriptions of images.

\subsection{Graph Linearization}
Before we move onto the second stage of SGRAM, we need to linearize AMR graphs into a sequence since the Transformer-based language model we use is the seq2seq model. Inspired by previous studies \cite{konstas-etal-2017-neural, cai-lam-2019-core, bevilacqua-etal-2021-one}, we adopt two graph traversal algorithms (DFS, BFS), and one tree traversal algorithm (in-order) to seek which graph traversal algorithm for graph linearization is more suitable for scene graph parsing. We take in-order traversal even though in-order performed poorly in the previous work \cite{Zhang2019AMRPA} since sorting in in-order traversal is similar to a form of sentence. 

\paragraph{Depth-First Search (DFS)} As mentioned in \cite{bevilacqua-etal-2021-one}, DFS is closely related to the linearized natural language syntactic trees. DFS starts at a root node and explores as deep as possible along each branch before moving on to next branch. For example, the AMR graph of region description, \textit{Golden retriever standing in the snow} in Figure \ref{fig:fig1_rev1} is linearized as: \textit{``(z0 / stand-01 :ARG1 (z1 / retriever :mod (z2 / gold)) :ARG2 (z3 / snow))"} where \textit{z} with numbers are special tokens to handle co-referring nodes generated from the AMR parsing model. Unlike SPRING model, we do not remove slash token and linearize the AMR graphs with DFS as it is. Thus, the root node ``\textit{(z0 / stand-01)}" comes first and explores as deep as possible to the left side of the root which is ``\textit{:ARG1 (z1 / retriever :mod (z2 / gold))}", then ``\textit{:ARG2 (z3 / snow))}" comes right after.

\paragraph{Breadth-First Search (BFS)} starts at a root node like DFS but explores all nodes at the present depth prior to moving on nodes at the next depth level. Following core semantic principle \cite{cai-lam-2019-core} which assumes core semantic nodes stay closely to the root, BFS algorithm can explore the most important nodes of a sentence first. So that the AMR graph of region description in Figure \ref{fig:fig1_rev1} is linearized as: ``\textit{(z0 / stand-01) :ARG1 (z1 / retriever) :ARG2 (z3 / snow) :mod (z2 / gold)}". Here, we remove the existing parentheses of AMR graphs generated from the AMR parsing model and add new parentheses to indicate each node. As shown the example, core semantics of the region description (i.e. ``\textit{(z0 / stand-01)}", ``\textit{(z1 / retriever)}", and ``\textit{(z3 / snow)}") are sorted first.

\paragraph{In-order traversal} follows the Left Root Right policy, which means a left subtree of a root node is traversed first, then the root node, and then the right subtree of the root node is traversed. We make the AMR graph as same as the region description by aligning the nodes and edges toward the root node according to the in-order traversal. So the AMR graph of region description in Figure \ref{fig:fig1_rev1} is linearizaed as: \textit{``(z2 / gold) :mod (z1 / retriever) :ARG1 (z0 / stand-01) :ARG2 (z3 / snow)"}. Here, we add new parentheses on nodes to indicate each node as same as BFS, and reverse the edges to make toward the root node as well.

\subsection{AMR-to-SG}

We take T5 model \cite{2020t5} as the Transformer-based language model to extend to generate scene graphs for the AMR graphs. T5 (Text-to-Text Transfer Transformer) is an encoder-decoder unified framework pre-trained on a multi-task mixture of unsupervised and supervised tasks. A wide range of natural language processing tasks such as translation, classification, and question answering are cast as feeding the model a text as input and training it to generate a target text. T5 comes in different sizes as follows: T5-small, T5-base, and T5-large. In particular, we use T5-base model for training our framework, which has 12 encoder and decoder layers and nearly 220M parameters. In addition to the T5 model we use, different size of T5 models and other models such as BART can be substituted for the use as well. 

For fine-tuning, the T5-base model takes linearized AMR graph as input sequence, so that the tokens of the linearized AMR graph for DFS are: \{\textit{`(z0/stand-01', `:ARG1', `(z1/retriever', `:mod', `(z2/gold))', `:ARG2', `(z3/snow))'}\}. For BFS, \{\textit{`(z0/stand-01)', `:ARG1', `(z1/retriever)', `:ARG2', `(z3/snow)', `:mod', `(z2/gold)'}\}. In-order will be same format as BFS but have different order. Then, those tokens are mapped into a task-specific output sequence in a form of \{\textit{(objects), (attribute-object), (object-relationship-object)}\}.
 
\begin{table}[]
\centering
\scalebox{1.0}{
\begin{tabular}{l|c}
\Xhline{2\arrayrulewidth}
    Scene graph parser model & F-score \\
    \hline
    Stanford \cite{schuster-etal-2015-generating} & 0.3549 \\
    SPICE \cite{anderson_spice_ECCV} & 0.4469 \\
    CDP \cite{wang-etal-2018-scene} & 0.4967 \\
    AG \cite{sgp_martin_neurips2019} & 0.5221 \\
    AG* \cite{sgp_martin_neurips2019} & 0.5750 \\
    \hline
    \textbf{Ours (SGRAM)} & \textbf{0.6128} \\
\Xhline{2\arrayrulewidth}
\end{tabular}}
\caption{F-score results of scene graph parsing comparison between existing parsers and SGRAM with variants (DFS, BFS, and In-order) on the subset of Visual Genome \cite{krishnavisualgenome}. CDP and AG are abbreviations of Customized Dependency Parser and Attention Graph, respectively. AG* is the model with limited number of words excluding ``\textit{a, an, the, and}" in the region descriptions.} 
\label{table:AMR_to_sg}
\end{table}

\section{Experiments}
\subsection{Datasets}

For fair comparisons with the existing parser models, we train and validate our framework with the subset of Visual Genome \cite{krishnavisualgenome} dataset used in the previous study \cite{wang-etal-2018-scene}. Each image in VG contains a number of regions, and each region is annotated with both a region description and a region scene graph. The training set is the intersection of the Visual Genome and MS COCO train2014 set, which contains 34,027 images with 1,070,145 regions. The evaluation set is the intersection of VG and MS COCO val2014 set, which contains 17,471 images with 547,795 regions. We follow the same preprocess steps as in \cite{wang-etal-2018-scene} for setting train/evaluation splits. In addition, there are some objects along with their attributes and relationships with other objects that are not related to region descriptions of images. For example, a region description, ``\textit{A person holding on umbrella}" contains an object ``\textit{bus}" with an attribute ``\textit{red}". Thus, we remove those not related to region descriptions in the training set.

\subsection{Evaluation}

To evaluate parsed scene graphs from AMR of region descriptions with the ground truth region scene graphs, we use SPICE metric \cite{anderson_spice_ECCV} which calculates a F-score over tuples as following equation: $F_{1}(g, r)=\frac{2 \cdot P(g, r) \cdot R(g, r)}{P(g, r) + R(g, r)}$ where \textit{P} and \textit{R} are precision and recall, respectively, and \textit{g} and \textit{r} are generated tuples and ground truth tuples of scene graph, respectively (refer Section 3.2 of \cite{anderson_spice_ECCV} for more details). As mentioned in \cite{wang-etal-2018-scene}, there is an issue that a node in one graph could be matched to several nodes in the other when SPICE calculates the F-score. Thus, following previous works, we enforce one-to-one matching while calculating the F-score and report the average F-score for all regions.

\begin{table}[]
\centering
\scalebox{1.0}{
\begin{tabular}{l|c}
\Xhline{2\arrayrulewidth}
    Model & F-score \\
    \hline
    \textbf{AMR2.0} & \\
    SGRAM (DFS) & \textbf{0.6128} \\
    SGRAM (BFS) & 0.6118\\
    SGRAM (In-order) & 0.6118 \\
    \hline
    \textbf{AMR3.0} & \\
    SGRAM (DFS) & 0.6114 \\
    SGRAM (BFS) & 0.6113 \\
    SGRAM (In-order) & 0.6110 \\
\Xhline{2\arrayrulewidth}
\end{tabular}}
\caption{F-score results of scene graph parsing comparison between variants of SGRAM (DFS, BFS, and In-order) on the subset of Visual Genome \cite{krishnavisualgenome}. AMR 2.0 and AMR 3.0 indicate the AMR parsers we use are pre-trained on AMR 2.0 or AMR 3.0.} 
\label{table:AMR2_0_3_0}
\end{table}

\subsection{Implementation details}

For our experiments, we implement our framework based on Pytorch. All of our models including DFS, BFS, and In-order graph linearization techniques use the same model hyperparameters as T5-base defined in Huggingface Transformer library, such as tokenizer, number of encoder and decoder layers.
Our final selection of hyperparameter settings from some trials of tuning is as follows. Our models are trained for $5$ epochs using cross-entropy with setting a fixed seed, batch size of $16$, maximum sequence length $1024$. Gradient accumulation is set to $1.0$. We set learning rate to $5 \times 10^{-5}$ with constant learning rate scheduler and weight decay $1 \times 10^{-4}$, and use AdaFactor \cite{Shazeer2018AdafactorAL} for optimizer. Beam size and dropout are set to $1$ and $0.1$, respectively. Training SGRAM framework takes about 21 hours on AMR 2.0 and AMR 3.0 using two Tesla V100 GPUs with 32 GB graphic memory.

\begin{table*}[t]
\centering
\begin{tabular}{l|ccc|cccc}
\Xhline{2\arrayrulewidth}
\multirow{2}{*}{Model} & \multicolumn{3}{c|}{\textbf{Development}} & \multicolumn{3}{c}{\textbf{Test}} \\
 & R@5 & R@10 & Med. Rank & R@5 & R@10 & Med. Rank\\
\hline
Stanford ~\cite{schuster-etal-2015-generating} & 33.82\% & 45.58\% & 6 & 34.96\% & 45.68\% & 5 \\
CDP ~\cite{wang-etal-2018-scene} & 36.69\% & 49.41\% & \textbf{4} & 36.70\% & 49.37\% & 5 \\
\hline
\textbf{AMR2.0} &  &  &  &  &  &  \\
SGRAM (DFS) & \textbf{37.41}\% & 50.51\% & \textbf{4} & \textbf{37.39}\% & \textbf{49.40}\% & \textbf{4} \\
SGRAM (BFS) & 37.16\% & 50.28\% & \textbf{4} & 36.84\% & 48.96\% & \textbf{4} \\
SGRAM (In-order) & 37.28\% & 50.29\% & \textbf{4} & 37.09\% & 49.32\% & \textbf{4} \\ 
\hline
\textbf{AMR3.0} &  &  &  &  &  &  \\
SGRAM (DFS) & 37.26\% & \textbf{50.61}\% & \textbf{4} & 37.21\% & 49.28\% & \textbf{4} \\ 
SGRAM (BFS) & 36.92\% & 50.17\% & \textbf{4} & 37.01\% & 48.99\% & \textbf{4} \\ 
SGRAM (In-order) & 37.16\% & 50.38\% & \textbf{4} & 37.09\% & 49.32\% & \textbf{4} \\
\Xhline{2\arrayrulewidth}
\end{tabular}
\caption{Image retrieval performance results comparison between existing parsers and SGRAM with variants on evaluation metric (Recall@k \{$5$, $10$\} and Median Rank). Stanford and CDP are abbreviation of Stanford scene graph parser \cite{schuster-etal-2015-generating} and Customized Dependency Parser \cite{wang-etal-2018-scene}, respectively.}
\label{table:img_ret}
\end{table*}

\begin{figure*}[]
\centering
    \includegraphics[width=0.9\textwidth]{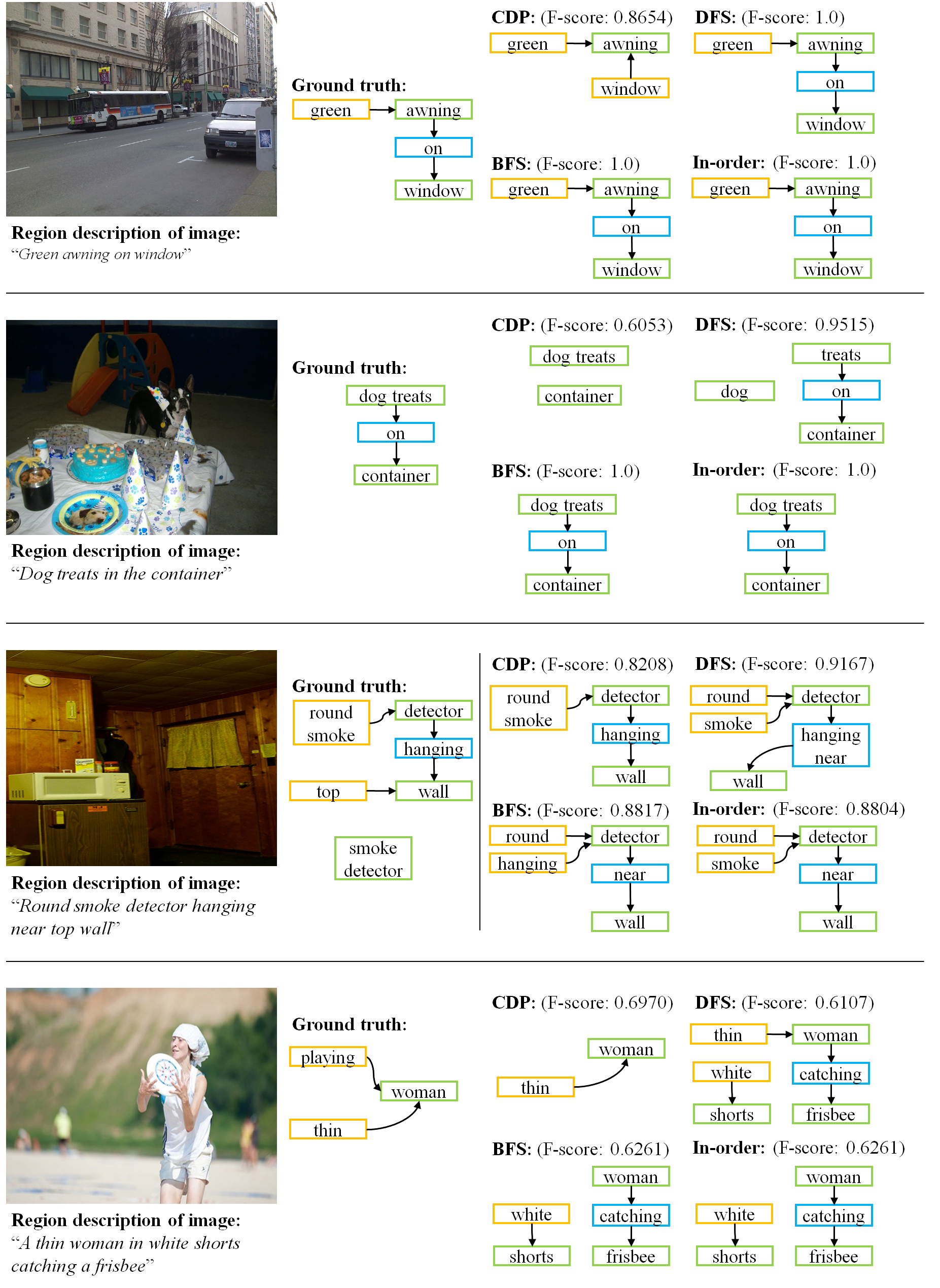}
\caption{Qualitative analysis on effectiveness of AMR in the evaluation set. Here, we visualize the outputs of CDP \cite{wang-etal-2018-scene} and SGRAM with various graph linearization techniques (DFS, BFS, and In-order). The boxes with green color indicate objects, the boxes colored in orange for attributes, and the boxes colored in sky-blue for relationships between objects. The edges are the connections between objects, attributes, and relationships.}
\label{fig:qualitative}
\end{figure*}
    
\subsection{Quantitative Analysis}
\paragraph{Comparison with previous methods} We report in Table \ref{table:AMR_to_sg}, the comparison results between SGRAM and previous works. As shown, our framework SGRAM achieves higher performance than existing dependency parsing models such as Stanford scene graph parser, SPICE, and CDP by a significant margin, 25.79\%, 16.59\% and 11.61\%, respectively. Furthermore, SGRAM achieve an F-score of 0.6128, which outperforms the the previous state-of-the-art model (AG*) \cite{sgp_martin_neurips2019} by 3.78\%. AG* is simple but efficient modification of AG model with input excluding ``\textit{a, an, the, and}" words in the region descriptions. Despite this measure, the performance of AG* remains poor compared to SGRAM. Based on the overall results, we confirm that AMR captures better high-level abstract semantics of textual descriptions than dependency parsing.

\paragraph{Analysis with graph linearization} As far as we consider linearization techniques, we show each linearization techniques with utilizing the AMR parser pre-trained on AMR 2.0 and 3.0. SGRAM (DFS) in both AMR 2.0 and 3.0 shows slightly better scene graph parsing performance than SGRAM (BFS) and SGRAM (In-order) as shown in Figure \ref{table:AMR2_0_3_0}. This may be because DFS is closely related to natural language synthetic trees. Furthermore, in the following qualitative analysis, SGRAM using BFS and in-order catches semantics of region descriptions well even if the performances are slightly lower. In consequence, we prove that scene graph parsing using AMR is robustly performed than previous methods regardless of whether AMR 2.0, AMR 3.0 or any graph traversal algorithm is used.

\subsection{Qualitative Analysis}
Figure \ref{fig:qualitative} provides the qualitative analysis on effectiveness of AMR. We represent region descriptions of image with the generated scene graphs of CDP \cite{wang-etal-2018-scene} and SGRAM (DFS, BFS, and In-order) with color boxes indicating green as nodes, orange as attribute, and sky-blue as relationships between objects, and also illustrate images to help understand the examples.
In the example on the first row, CDP does not capture the relationship between ``\textit{awning}" and ``\textit{widnow}", which is ``\textit{awning - on - window}" because ``\textit{window}" in region description is taken as an attribute. On the other hand, all SGRAM with various linearization techniques obtain an F-score of $1.0$ by capturing all the semantics of the region description. In the example on the second row, CDP only catches objects, ``\textit{dog treat}" and ``\textit{container}", whereas all of SGRAM with graph linearization captures the relationship ``\textit{dog treats - on - container}" with the objects. However, DFS captures ``\textit{dog treats}" as two objects, ``\textit{dog}" and ``\textit{treats}". In the third row example, although the attribute of ``\textit{detector}" is ``\textit{round smoke}" in the attribute tuple of ground truth, both CDP and SGRAM shows more than 0.8 F-score performance. CDP captures ``\textit{round smoke}" correctly, and SGP with DFS and In-order captures the attribute ``\textit{round smoke}" separately, ``\textit{round}" and ``\textit{smoke}". BFS catches ``\textit{hanging}" as the attribute of ``\textit{detector}", which can be correct even though not in the ground truth tuple. We consider that ``\textit{detector - smoke}" and ``\textit{detector - round}" like the output of DFS and In-order are more accurate attributes than ``\textit{round smoke}". In addition, all models except SGRAM (DFS) take the relationship between ``\textit{detector}" and ``\textit{wall}" as ``\textit{hanging}" while SGRAM (DFS) took a more detailed relationship as ``\textit{hanging near}". On the fourth row example, the ground truth tuples have only ``\textit{woman}"-related tuples, no object tuples of ``\textit{shorts}" or ``\textit{frisbee}", and no attribute or relationship tuples for them. For that reason, SGRAM with linearization techniques obtained a lower F-score than CDP, although most of semantics in the region description such as ``\textit{catching}", ``\textit{shorts - white}", and ``\textit{woman - catching - frisbee}" were captured well.

In summary, we demonstrate that AMR has much better F-score performance than CDP and captures detailed semantics of the region descriptions such as ``\textit{round}", ``\textit{smoke}" and ``\textit{hanging near}". We also confirm that there are cases in which performance is not achieved well due to poor ground truth labels even though AMR picked out semantics well.

\subsection{Effectiveness of AMR in Image Retrieval}

We apply the effectiveness of SGRAM with various graph linearization to the image retrieval task, which is one of scene graph downstream tasks. For performance comparison with existing parsers, the development and test set of the image retrieval dataset \cite{schuster-etal-2015-generating} are used. The development and test sets of the dataset have $454$ and $456$ images respectively, and all regions in an image have human-annotated region descriptions and region scene graphs. The average number of region descriptions for an image is $11.2$. We follow all experimental setups of \cite{wang-etal-2018-scene}. As mentioned in \cite{wang-etal-2018-scene}, we also could not obtain or reproduce the CRF model used in \cite{schuster-etal-2015-generating}, so we use F-score similarity instead of the likelihood of the maximum a posteriori CRF solution when we rank images. The major difference from previous works is that we construct AMR graphs from region descriptions and use those as a query, not the regional descriptions themselves. the queries are linearized as same as we propose in SGRAM framework. Furthermore, We make an attempt to report the image retrieval performance of customized attention graph model \cite{sgp_martin_neurips2019} by reproducing but the source code of has not been released.

The results are shown in Table \ref{table:img_ret}. In overall, our SGRAM performs 0.66\% better on average than the existing scene graph parsers in the development and test sets. In particular, SGRAM (DFS) models show better performance than SGRAM using BFS or In-order. To sum up, AMR shows good performance in image retrieval by catching semantics of sentences better than dependency parsing. We believe that this can be affected to other vision tasks that require high-level semantics such as object grounding \cite{Yi2022IncrementalOG} and multi-modal pre-training \cite{Li2021UNIMOTU}. 

\section{Conclusion}
In this paper we design a simple yet effective two-stage scene graph parsing framework utilizing abstract meaning representation, SGRAM (Scene GRaph parsing via Abstract Meaning representation). By applying SGRAM to textual scene graph parsing, we confirm that the performance of our framework outperforms the performance of the previous state-of-the-art model as well as the existing dependency parsing-based model. Furthermore, we demonstrate that SGRAM performs textual scene graph parsing robustly regardless of AMR 2.0, AMR 3.0, and graph linearization techniques. We perform the image retrieval task that requires high-level reasoning to confirm the expressive power of AMR graphs, the result of the first stage in SGRAM. Similarly, we show competitive performance when comparing the performance with comparative methods, confirming that AMR graphs can be applied to a variety of downstream tasks. As a future work, we expect that AMR will be an effective semantic representation in video domain as well as image domain, and suggest to perform textual scene graph parsing using the AMR parsing model with better performance and investigate an adapter-based Transformer-based language model to make training simpler and faster.

% Use \bibliography{yourbibfile} instead or the References section will not appear in your paper
%\bibliography{arxiv_wschoibbl}

\end{document}